%% file: paper.tex
\documentclass{article} % For LaTeX2e
\usepackage{iclr2016_conference,times}
\usepackage{hyperref}
\usepackage{wrapfig}
\usepackage{url}
\usepackage{tikz}
\usepackage{xcolor,colortbl}
\usepackage{amssymb}
\usepackage{amsmath}
\usepackage{float}
\usepackage{afterpage}
\iclrfinalcopy
\title{Neural Random-Access Machines}

\author{Karol Kurach\thanks{Equal contribution.}\:\:\:\& Marcin Andrychowicz$^*$ \& Ilya Sutskever \\
Google\\
\texttt{\{kkurach,marcina,ilyasu\}@google.com} \\
}

\newcommand{\mem}{\mathcal{M}}
\newcommand{\ZM}{\{0,1,\ldots,M-1\}}

\begin{document}

\maketitle

\input{abstract}
\input{intro}
\input{related_work}

\input{model}
\input{experiments}

\input{conclusions}

\newpage
\bibliographystyle{iclr2016_conference}
\bibliography{paper}
\newpage
\appendix
\input{task_descriptions}

\newpage
\input{appendix_curriculum}
\newpage
\input{appendix_circuits}

\end{document}

%% file: abstract.tex
\begin{abstract}

In this paper, we propose and investigate a new neural network architecture called Neural Random Access Machine.
It can manipulate and dereference pointers to an external variable-size random-access memory.
The model is trained from pure input-output examples using backpropagation.

We evaluate the new model on a number of simple algorithmic tasks whose solutions require pointer manipulation
and dereferencing.  Our results show that the proposed model can learn
to solve algorithmic tasks of such type and is capable of operating on simple data structures like linked-lists and binary trees.
For easier tasks, the learned solutions generalize to sequences of arbitrary length. 
Moreover, memory access during inference can be done in a constant time under some assumptions.

\end{abstract}

%% file: intro.tex
\section{Introduction}

Deep learning is successful for two reasons. First, deep neural
networks are able to represent the ``right'' kind of functions;
second, deep neural networks are trainable.  Deep neural networks can
be potentially improved if they get deeper and have fewer
parameters, while maintaining trainability. By doing so, we move
closer towards a practical implementation of Solomonoff induction
\citep{solomonoff}. The first model that we know of that attempted to
train extremely deep networks with a large memory and few
parameters is the Neural Turing Machine (NTM) \citep{ntm} --- a computationally
universal deep neural network that is trainable with backpropagation.
Other models with this property include variants of
Stack-Augmented recurrent neural networks \citep{stack-rnn,stack2},
and the Grid-LSTM \citep{grid}---of which the Grid-LSTM has achieved
the greatest success on both synthetic and real tasks. The key
characteristic of these models is that their depth, the size of their
short term memory, and their number of parameters are no longer
confounded and can be altered independently --- which stands in
contrast to models like the LSTM \citep{lstm}, whose number of
parameters grows quadratically with the size of their short term
memory.

A fundamental operation of modern computers is pointer manipulation
and dereferencing. In this work, we investigate a model class that we name the Neural
Random-Access Machine (NRAM), which is a neural network that has, as
primitive operations, the ability to manipulate, store in memory,
and dereference pointers into its working memory. By providing our
model with dereferencing as a primitive, it becomes possible to train
models on problems whose solutions require pointer manipulation and
chasing. Although all computationally universal neural networks
are equivalent, which means that the NRAM model does not have a
representational advantage over other models if they are given a
sufficient number of computational steps, in practice, the number of
timesteps that a given model has is highly limited, as extremely deep
models are very difficult to train. As a result, the model's 
core primitives have a strong effect on the set of functions that can be
feasibly learned in practice, similarly to the way in which the choice
of a programming language strongly affects the functions that can be
implemented with an extremely small amount of code.

Finally, the usefulness of computationally-universal neural networks
depends entirely on the ability of backpropagation to find good
settings of their parameters. Indeed, it is trivial to define the
``optimal'' hypothesis class \citep{solomonoff}, but the problem of
finding the best (or even a good) function in that class is
intractable. Our work puts the backpropagation algorithm to another
test, where the model is extremely deep and intricate.

In our experiments, we evaluate our model on several algorithmic
problems whose solutions required pointer manipulation and chasing.
These problems include algorithms on a linked-list and a binary tree.
While we were able to achieve encouraging results on these problems, we
found that standard optimization algorithms struggle with these
extremely deep and nonlinear models. We believe that advances in
optimization methods will likely lead to better results.

%% file: related_work.tex
\section{Related work}

There has been a significant interest in the problem of learning
algorithms in the past few years.  The most relevant recent paper
is Neural Turing Machines (NTMs) \citep{ntm}.  It was the
first paper to explicitly suggest the notion that it is worth
training a computationally universal neural network, and achieved
encouraging results.

A follow-up model that had the goal of learning algorithms was the
Stack-Augmented Recurrent Neural Network \citep{stack-rnn}
This work demonstrated that the
Stack-Augmented RNN can generalize to long problem instances from
short problem instances.  A related model is the Reinforcement
Learning Neural Turing Machine \citep{rl-ntm}, which attempted to use
reinforcement learning techniques to train a discrete-continuous
hybrid model.%, which has been successful on simple

The memory network \citep{memnet1} is an early model that attempted  
to explicitly separate the memory from computation in a neural network model.
The followup work of \cite{memnet2} combined the memory network with the soft
attention mechanism, which allowed it to be trained with less supervision.

The Grid-LSTM \citep{grid} is a highly interesting extension of LSTM,
which allows to use LSTM cells for both deep and sequential
computation.  It achieves excellent results on both synthetic,
algorithmic problems and on real tasks, such as language modelling,
machine translation, and object recognition.

The Pointer Network \citep{vinyals2015pointer} is somewhat different
from the above models in that it does not have a writable memory ---
it is more similar to the attention model of \citet{bahdanau2014} in
this regard.  Despite not having a memory, this model was able to solve
a number of difficult algorithmic problems that include the convex hull
and the approximate 2D travelling salesman problem (TSP).    

Finally, it is important to mention the attention model of
\citet{bahdanau2014}.  Although this work is not explicitly aimed at
learning algorithms, it is by far the most practical model that has an
``algorithmic bent''.  Indeed, this model has proven to be highly
versatile, and variants of this model have achieved state-of-the-art
results on machine translation \citep{thang-mt}, speech recognition
\citep{las}, and syntactic parsing \citep{parsing}, without the use of
almost any domain-specific tuning.

%% file: model.tex
\section{Model}

In this section we describe the NRAM model.
We start with a description of the simplified version of our model which does not use an external memory and then
explain how to augment it with a variable-size random-access memory.
The core part of the model is a neural controller, which acts as a ``processor''.
The controller can be a feedforward neural network or an LSTM, and it is the only trainable part of the model.

The model contains $R$ registers, each of which holds an integer value.
To make our model trainable with gradient descent, we made it fully differentiable.
Hence, each register represents an integer value with a 
distribution over the set $\ZM$, for some constant $M$.
We do not assume that these distributions have any special form --- they are simply stored as vectors $p \in \mathbb{R}^M$
satisfying $p_i \ge 0$ and $\sum_i p_i = 1$.
The controller does not have direct access to the registers; it can interact with them
using a number of prespecified \emph{modules} (gates), such as integer addition
or equality test.
Let's denote the modules $m_1,m_2,\ldots,m_Q$, where
each module is a function:
$$m_i : \ZM \times \ZM \rightarrow \ZM.$$

On a high level, the model performs a sequence of timesteps, each of which consists of the following substeps:
\begin{enumerate}
 \item The controller gets some inputs depending on the values of the registers
 (the controller's inputs are described in Sec.~\ref{sec:disc}).
 \item The controller updates its internal state (if the controller is an LSTM).
 \item The controller outputs the description of a ``fuzzy circuit'' with inputs $r_1,\ldots,r_R$,
 gates $m_1, \ldots, m_Q$ and $R$ outputs.
 \item The values of the registers are overwritten with the outputs of the circuit.
\end{enumerate}

More precisely, each circuit is created as follows.
The inputs for the module $m_i$ are chosen by the controller from the set
$\{r_1,\ldots,r_R,o_1,\ldots,o_{i-1}\}$, where:
\begin{itemize}
 \item $r_j$ is the value stored in the $j$-th register at the current timestep, and
 \item $o_j$ is the output of the module $m_j$ at the current timestep.  
\end{itemize}

Hence, for each $1 \le i \le Q$ the controller chooses weighted averages % (with coefficients summing to $1$) 
of the values
$\{r_1,\ldots,r_R,o_1,\ldots,o_{i-1}\}$ which are given as inputs to the
module.  Therefore,
\begin{equation}\label{eq:mod}
o_i = m_i \left( (r_1,\ldots,r_R,o_1,\ldots,o_{i-1})^T \mathbf{softmax}(a_i),\,\, (r_1,\ldots,r_R,o_1,\ldots,o_{i-1})^T \mathbf{softmax}(b_i) \right),
\end{equation}
where the vectors $a_i,b_i \in \mathbb{R}^{R+i-1}$ are produced by the controller (Fig.~\ref{fig:inputs}).

\usetikzlibrary{decorations.pathreplacing,calc,matrix,circuits.logic.US,positioning,matrix,automata}

\begin{figure}[h]
\begin{center}
\begin{tikzpicture}[circuit logic US,
                    large circuit symbols,
                    every circuit symbol/.style={fill=white,draw, logic gate input sep=2mm}
                    ]
\matrix (R)[matrix of nodes,row sep=-\pgflinewidth,nodes={rectangle,draw},every node/.style={anchor=base,text depth=.5ex,text height=1.5ex,text width=1.5em,rectangle,draw},align=center] at (-4,0)
  {
    $r_1$  &
    $\ldots$  &
    $r_R$  &
    $o_1$  &
    $\ldots$  &
    $o_{i-1}$ \\
  };
  \node[above left=-0.1cm and -2.0cm of R] {registers};
  \node[above left=0.0cm and  -4.7cm of R, text width=2cm,align=center] {outputs of previous modules};

 % \node (I) at (-4,0) {$(R_1,\ldots,r_R,o_1,\ldots,o_{i-1})$}

 \node[draw,rectangle] (LSTM) at (-3,-1.5) {LSTM};                    
  
 \node[or gate,draw,point right]  (m1)  at (2,0.5) {$\langle\cdot,\cdot\rangle$};
 \node[or gate,draw,point right]  (m2)  at (2,-0.5) {$\langle\cdot,\cdot\rangle$};
 \node[or gate,draw,point right]  (s1)  at (0,-1) {s-m};
 \node[or gate,draw,point right]  (s2)  at (0,-2) {s-m};

 \node[or gate,draw,point right]  (m3)  at (4,0) {$m_i$};
 
 \draw[->] (s1.output)  to[bend right=0] (m1.input 2);
 \draw[->] (s2.output)  to[bend right=0] (m2.input 2);
 \draw[->] (m1.output)  to[bend right=20] (m3.input 1);
 \draw[->] (m2.output)  to[bend right=-20] (m3.input 2);
 \node[right=1cm of m3.output] (o) {$o_i$};
 \draw[->] (m3.output) to[bend right=0] (o);
 
 \draw[->] (R.east) to[bend right=0] (m1.input 1);
 \draw[->] (R.east) to[bend right=0] (m2.input 1);
 \draw[->] (LSTM.east) -- (s1.west) node[midway,yshift=0.2cm] {$a_i$};
 \draw[->] (LSTM.east) -- (s2.west) node[midway,yshift=-0.2cm] {$b_i$};
 
\end{tikzpicture}
\end{center}
\caption{The execution of the module $m_i$. Gates \emph{s-m} represent the softmax function and $\langle\cdot,\cdot\rangle$ denotes inner product. See Eq.~\ref{eq:mod} for details.
}\label{fig:inputs}
\end{figure}
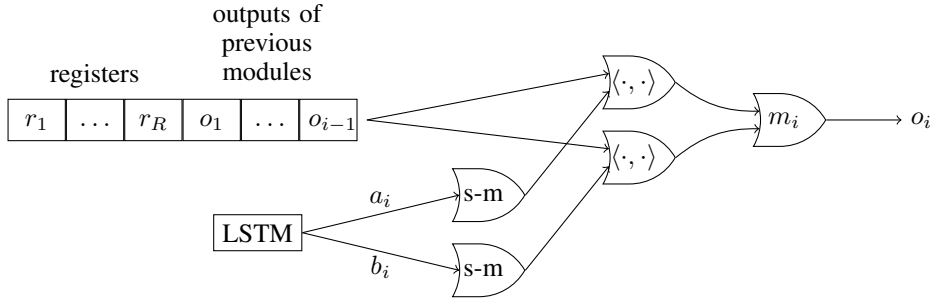

Recall that the variables $r_j$ represent probability distributions
and therefore the inputs to $m_i$, being weighted averages of
probability distributions, are also probability distributions.  Thus,
as the modules $m_i$ are originally defined for integer inputs and
outputs, we must extend their domain to probability distributions as
inputs, which can be done in a natural way (and make their output also
be a probability distribution):
\begin{equation}\label{eq:fuzzy}
\forall_{0 \le c < M}\,\,\, \mathbb{P} \left( m_i(A,B) = c \right) = \sum_{0 \le a,b < M} \mathbb{P}(A=a) \mathbb{P}(B=b) [m_i(a,b)=c].
\end{equation}

After the modules have produced their outputs, the controller decides
which of the values $\{r_1,\ldots,r_R,o_1,\ldots,o_Q\}$ should be stored in the registers.
In detail, the controller outputs the vectors $c_i \in \mathbb{R}^{R+Q}$ for $1 \le i \le R$
and the values of the registers are updated (simultaneously) using the formula:
\begin{equation}\label{eq:reg}
r_i := (r_1,\ldots,r_R,o_1,\ldots,o_Q)^T \mathbf{softmax}(c_i).
\end{equation}

\subsection{Controller's inputs}\label{sec:disc}

Recall
that at the beginning of each timestep the controller receives some
inputs, and it is an important design decision to decide where should
these inputs come from.  A naive approach is to use the values of the
registers as inputs to the controller.  However, the values of the registers are
probability distributions and are stored as vectors $p \in
\mathbb{R}^M$.  If the entire distributions were given as inputs to
the controller then the number of the model's parameters would depend
on $M$.
This would be undesirable because, as will be explained in the next section, the value $M$
is linked to the size of an external random-access memory tape
and hence it would prevent the model from generalizing to different memory sizes.

Hence, for each $1 \le i \le R$ the
controller receives, as input, only one scalar from each register,
namely $\mathbb{P}(r_i = 0)$ --- the probability that the value in the
register is equal $0$.
This solution has an additional advantage, namely it limits the amount
of information available to the controller and forces it to
rely on the modules instead of trying to solve the problem on its own.
Notice that this information is sufficient to get the exact value of $r_i$ if
$r_i \in \{0,1\}$, which is the case whenever $r_i$ is an output of a ,,boolean'' module, e.g.
the inequality test module $m_i(a,b)=[a<b]$.

\subsection{Memory tape}\label{sec:memory}

One could use the model described so far for learning
sequence-to-sequence transformations by initializing the registers
with the input sequence, and training the model to produce the desired output
sequence in its registers after a given number of
timesteps.  The disadvantage of such model is that it would be
completely unable to generalize to longer sequences, because the length of the sequence
that the model can process is equal to the number of its registers, which is constant.

Therefore, we extend the model with a variable-size memory tape, which
consists of $M$ memory cells, each of which stores a distribution over
the set $\ZM$.  Notice that each distribution stored in a memory cell
or a register can be interpreted as a fuzzy address in the memory
and used as a fuzzy pointer. We will hence identify integers in the
set $\ZM$ with pointers to the memory.
Therefore, the value in each memory cell may be interpreted as an integer or as a pointer.
The exact state of the memory
can be described by a matrix $\mem \in \mathbb{R}^M_M$, where the
value $\mem_{i,j}$ is the probability that the $i$-th cell holds the
value $j$.

The model interacts with the memory tape solely using two special modules:
\begin{itemize}

 \item \texttt{READ} module: this module takes as the input
 a pointer\footnote{Formally each module takes two arguments. In this case the
 second argument is simply ignored.} and returns the value stored under the
 given address in the memory.  This operation is extended to fuzzy pointers
 similarly to Eq.~\ref{eq:fuzzy}.  More precisely, if $p$ is a vector
 representing the probability distribution of the input (i.e. $p_i$ is the
 probability that the input pointer points to the $i$-th cell) then the module
 returns the value $\mem^T p$.

\item \texttt{WRITE} module: this module takes as the input a pointer $p$ and
a value $a$ and stores the value $a$ under the address $p$ in the memory.  The
fuzzy form of the operation can be effectively expressed using matrix
operations \footnote{The exact formula is $\mem := (J-p)J^T \cdot \mem + pa^T$,
where $J$ denotes a (column) vector consisting of $M$ ones and $\cdot$ denotes
coordinate-wise multiplication.}.

\end{itemize}

The full architecture of the NRAM model is presented on Fig.~\ref{fig:nram}

\usetikzlibrary{decorations.pathreplacing,calc,matrix,circuits.logic.US,positioning}

\begin{figure}[h]
\begin{center}
\begin{tikzpicture}[circuit logic US,
                    tiny circuit symbols,
                    every circuit symbol/.style={fill=white,draw, logic gate input sep=2mm}
                    ]
                    
\matrix (R)[matrix of nodes,row sep=-\pgflinewidth,nodes={rectangle,draw}] at (-5,0)
  {
    $r_1$  \\
    $r_2$  \\
    $r_3$  \\
    $r_4$  \\
  };
\node[rotate=90, left=0.3cm of R-1-1] {registers};
  
\node[or gate,draw,point right]  (m1)  at (-3,0.5) {$m_1$};
\draw[->] (R-1-1.east) to[bend right=0] (m1.input 1);
\draw[->] (R-3-1.east) to[bend right=-20] (m1.input 2);

\node[or gate,draw,point right]  (m2)  at (-1,-0.5) {$m_2$};
\draw[->] (m1.output)  to[in=180, bend right] (m2.input 1);
\draw[->] (R-3-1.east) to[bend right=0] (m2.input 2);

\node[or gate,draw,point right]  (m3)  at (1,0.5) {$m_3$};
\draw[->] (m1.output)  to[bend right=0] (m3.input 1);
\draw[->] (m2.output)  to[in=180] (m3.input 2);

\draw[loosely dotted, line width=1pt] (2.5,0) -- (3.5,0);

\matrix (Rp)[matrix of nodes,row sep=-\pgflinewidth,nodes={rectangle,draw}] at (5,0)
  {
    $r_1$  \\
    $r_2$  \\
    $r_3$  \\
    $r_4$  \\
  };

\draw[->] (m3.output)  -- ([xshift=20]m3.output);    
  
\draw[->] ([xshift=-20]Rp-1-1.west)  -- (Rp-1-1);  
\draw[->] ([xshift=-20]Rp-2-1.west)  -- (Rp-2-1);  
\draw[->] (m2.output)  -- (Rp-3-1.west);  
\draw[->,bend right] (R-4-1.east)   to[bend right=10] (Rp-4-1.west);  

\draw [decoration={brace,amplitude=0.5em},decorate,thick]
 (R.north east) --  (Rp.north west);

\node[draw,rectangle] (LSTM) at (0,2) {LSTM};                    
\node[draw,rectangle] (finish) at (3,2) {finish?};
\draw[->,thick] (LSTM.east) to[in=90, out=0, looseness=4] (LSTM.north);
\draw[->,thick] (R-1-1.north) to[in=180, out=90] node[above] {binarized} (LSTM.west); 
\draw[->,thick] (LSTM) to (0,1.3); 
\draw[->,thick] (LSTM) to (finish);

\draw[step=0.5cm,black,xshift=0.25cm] (-5.5,-2.5) grid (5,-2);
\node at (0,-2.8) {memory tape};
\draw[->,dashed] (m1) to[bend left=30] ([yshift=-2.5cm]m1); 
\draw[<-,dashed] (m1) to[bend right=30] ([yshift=-2.5cm]m1); 
\draw[->,dashed] (m2) to[bend left=30] ([yshift=-1.5cm]m2); 
\draw[<-,dashed] (m2) to[bend right=30] ([yshift=-1.5cm]m2); 
\draw[->,dashed] (m3) to[bend left=30] ([yshift=-2.5cm]m3); 
\draw[<-,dashed] (m3) to[bend right=30] ([yshift=-2.5cm]m3); 

\draw [ultra thick, draw=black, fill=gray, opacity=0.2]
       (R.north east) -- (Rp.north west) -- ([yshift=-0.3cm]Rp.south west) -- ([yshift=-0.3cm]R.south east) -- cycle;
 
\end{tikzpicture}
\end{center}
\caption{One timestep of the NRAM architecture with $R=4$ registers.
The LSTM controller gets the ,,binarized'' values $r_1,r_2,\ldots$ stored in the registers as inputs
and outputs the description of the circuit in the grey box and the probability of
finishing the execution in the current timestep (See Sec.~\ref{sec:inputs} for more detail).
The weights of the solid thin connections are outputted by the controller.
The weights of the solid thick connections are trainable parameters of the model.
Some of the modules (i.e. \texttt{READ} and \texttt{WRITE})
may interact with the memory tape (dashed connections).
}\label{fig:nram}
\end{figure}
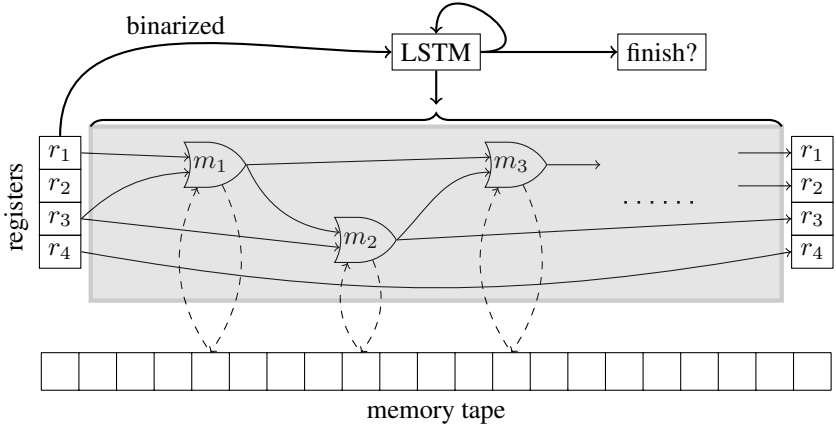

\subsection{Inputs and outputs handling}\label{sec:inputs}

The memory tape also serves as an input-output channel ---
the model's memory is initialized with the input sequence
and the model is expected to produce the output in the memory.
Moreover, we use a novel way of deciding how many timesteps should be executed.
After each timestep we let the controller decide whether it would like to continue
the execution or finish it, in which case the current state of the memory is treated as the output.

More precisely, after the timestep $t$ the controller outputs a scalar $f_t \in [0,1]$\footnote{In fact,
the controller outputs a scalar $x_i$ and $f_i=\mathbf{sigmoid}(x_i)$.}, which denotes
the willingness to finish the execution in the current timestep.
Therefore, the probability that the execution has not been finished before the timestep $t$ is
equal $\prod_{i=1}^{t-1} (1 - f_i)$,
and the probability that the output is produced exactly at the timestep $t$ is equal
$p_t = f_t \cdot \prod_{i=1}^{t-1} (1 - f_i)$.
There is also some maximal allowed number of timesteps $T$, which is
a hyperparameter.
The model is forced to produce output in the last step if it has
not done it yet, i.e. $p_T = 1 - \sum_{i=1}^{T-1} p_i$ regardless of the value
$f_T$.

Let $\mem^{(t)} \in \mathbb{R}^M_M$ denote the memory matrix after the timestep $t$, i.e.
$\mem^{(t)}_{i,j}$ is the probability that the $i$-th memory cell holds the value $j$ after the timestep $t$.
For an input-output pair $(x,y)$, where $x,y \in \ZM^M$ we define the
loss of the model as the expected negative log-likelihood of producing the correct
output, i.e., $-\sum_{t=1}^T \left( p_t \cdot \sum_{i=1}^M \log (\mem^{(t)}_{i,y_i}) \right)$
assuming that the memory was initialized with the sequence $x$\footnote{One could
also use the negative log-likelihood of the expected output, i.e.
$-\sum_{i=1}^M \log \left(\sum_{t=1}^T p_t \cdot \mem^{(t)}_{i,y_i} \right)$ as the loss
function.}.  
Moreover, for all problems we consider the output sequence is shorter than the memory.
Therefore, we compute the loss only over memory cells, which should contain the output.

\subsection{Discretization}\label{sec:discrete}

Computing the outputs of modules, represented as probability distributions,
is a computationally costly operation.
For example, computing the output of the \texttt{READ} module
takes $\Theta(M^2)$ time as it requires the multiplication of the matrix $\mem \in \mathbb{R}^M_M$
and the vector $p \in \mathbb{R}^M$.

One may however suspect (and we empirically verify this claim in
Sec.~\ref{sec:exp}) that the NRAM model naturally learns solutions in
which the distributions of intermediate values have very low entropy.
The argument for this hypothesis is that fuzziness in the
intermediate values would probably propagate to the output and cause a
higher value of the cost function.  To test this hypothesis we trained
the model and then used its \emph{discretized} version during
interference.  In the discretized version every module gets as inputs
the values from modules (or registers), which are the most probable to
produce the given input accordingly to the distribution outputted by
the controller.  More precisely, it corresponds to replacing the
function $\mathbf{softmax}$ in equations (\ref{eq:mod},\ref{eq:reg})
with the function returning the vector containing $1$ on the position
of the maximum value in the input and zeros on all other
positions.%\footnote{Draws can be handled arbitrarily.}.

Notice that in the discretized NRAM model each register and memory cell stores an integer from the set $\ZM$
and therefore all modules may be executed efficiently (assuming that the functions represented by the modules
can be efficiently computed).
In case of a feedforward controller and a small (e.g. $\le 20$) number of registers the interference can be accelerated even further.
Recall that the only inputs to the controller are binarized values of the register.
Therefore, instead of executing the controller one may simple precompute the
(discretized) controller's output for each configuration of the registers' binarized values.
Such algorithm would enjoy an extremely efficient implementation in machine code.

%% file: experiments.tex
\section{Experiments}\label{sec:exp}

\subsection{Training procedure}\label{sec:training}
\label{training_procedure}

The NRAM model is fully differentiable and we trained it using the
Adam optimization algorithm \citep{kingma2014adam} with the negative log-likelihood
cost function. Notice that we do not use any additional supervised
data (such as memory access traces) beyond pure input-output examples.

We used multilayer perceptrons (MLPs) with two hidden layers or LSTMs
with a hidden layer between input and LSTM cells as controllers.
The number of hidden units in each layer was equal.
The ReLu nonlinearity \citep{relu} was used in all experiments.

Below are some important techniques that we used in the training:

\paragraph{Curriculum learning}
As noticed in several papers \citep{curriculum,lte}, curriculum
learning is crucial for training deep networks on very complicated
problems. We followed the curriculum learning schedule from
\citet{lte} without any modifications. The details can be found in Appendix~\ref{appendix:curriculum}.

\paragraph{Gradient clipping}
  Notice that the depth of the unfolded execution is roughly a product of the
  number of timesteps and the number of modules.  Even for moderately small
  experiments (e.g. $14$ modules and $20$ timesteps) this value easily exceeds
  a few hundreds.  In networks of such depth, the gradients can often
  ``explode'' \citep{bengio}, what makes training by backpropagation much
  harder.  We noticed that the gradients w.r.t.~the intermediate values inside
  the backpropagation were so large, that they sometimes led to an overflow in
  single-precision floating-point arithmetic.  Therefore, we clipped the
  gradients w.r.t.~the activations, within the execution of the
  backpropagation algorithm. More precisely, each coordinate is separately
  cropped into the range $[-C_1,C_1]$ for some constant $C_1$.  Before updating
  parameters, we also globally rescale the whole gradient vector, so that its
  L2 norm is not bigger than some constant value $C_2$.

\paragraph{Noise} We added random Gaussian noise to the computed gradients
after the backpropagation step. The variance of this noise decays exponentially
during the training. The details can be found in \cite{neelakantan2015adding}.

\paragraph{Enforcing Distribution Constraints} For very deep networks, a small error in one place
  can propagate to a huge error in some other place. This was the case with our
  pointers: they are probability distributions over memory cells and they
  should sum up to 1. However, after a number of operations are applied, they can
  accumulate error as a result of inaccurate floating-point arithmetic.
  
  We have a special layer which is responsible for rescaling all values
  (multiplying by the inverse of their sum), to make sure they always represent
  a probability distribution. We add this layer to our model in a few critical
  places (eg. after the softmax operation)\footnote{We do not however
  backpropagate through these renormalizing operations, i.e. during the
  backward pass we simply assume that they are identities.}.
 
\paragraph{Entropy}
While searching for a solution, the network can fix the pointer
distribution on some particular value.  This is advantageous at the end
of training, because ideally we would like to be able to discretize the model.
However, if this happens at the begin of the training, it could force the
network to stay in a local minimum, with a small chance of moving the
probability mass to some other value. To address this problem, we encourage the
network to explore the space of solutions by adding an "entropy bonus", that
decreases over time. More precisely, for every distribution outputted by the controller, we subtract from the cost function
the entropy of the distribution multiplied by some coefficient, which decreases
exponentially during the training.

\paragraph{Limiting the values of logarithms}
There are two places in our model where the logarithms are computed
--- in the cost function and in the entropy computation.
Inputs to whose logarithms can be very small numbers, which may cause
very big values of the cost function or even overflows in
floating-point arithmetic.  To prevent this phenomenon we use
$\log(\max(x,\epsilon))$ instead of $\log(x)$ for some small
hyperparameter $\epsilon$ whenever a logarithm is computed.

\subsection{Tasks}\label{sec:tasks}

We now describe the tasks used in our experiments. For every task, the input is
given to the network in the memory tape, and the network's goal is to modify
the memory according to the task's specification. We allow the network to modify the original
input. The final error for a test case is computed as $\frac{c}{m}$, where $c$
is the number of correctly written cells, and $m$ represents the total number
of cells that should be modified.

Due to limited space, we describe the tasks only briefly here. The detailed memory layout of inputs and outputs can be found in the Appendix~\ref{appendix:tasks}.

\begin{enumerate} 

\item \textbf{Access} Given a value $k$ and an array $A$, return $A[k]$. 

\item \textbf{Increment} Given an array, increment all its elements by 1.

\item \textbf{Copy} Given an array and a pointer to the destination, copy all
elements from the array to the given location. 

\item \textbf{Reverse} Given an array and a pointer to the destination, copy
all elements from the array in reversed order. 

\item \textbf{Swap} Given two pointers $p$, $q$ and an array $A$, swap elements
$A[p]$ and $A[q]$. 

\item \textbf{Permutation} Given two arrays of $n$ elements: $P$ (contains
a permutation of numbers ${1,\ldots,n}$) and $A$ (contains random elements), permutate $A$
according to $P$. 

\item \textbf{ListK} Given a pointer to the head of a linked list and a number $k$,
find the value of the $k$-th element on the list. 

\item \textbf{ListSearch} Given a pointer to the head of a linked list and a value $v$ to find
return a pointer to the first node on the list with the value $v$.

\item \textbf{Merge}
Given pointers to $2$ sorted arrays $A$ and $B$, merge them.

\item \textbf{WalkBST} 
Given a pointer to the root of a Binary Search Tree, and a path to be
traversed (sequence of left/right steps), return the element at the end of the path. 

\end{enumerate}

\subsection{Modules}\label{sec:modules}

In all of our experiments we used the same sequence of $14$ modules:
 \texttt{READ} (described in Sec.~\ref{sec:memory}),
 $\text{\texttt{ZERO}}(a,b)=0$,
 $\text{\texttt{ONE}}(a,b)=1$,
 $\text{\texttt{TWO}}(a,b)=2$,
 $\text{\texttt{INC}}(a,b)=(a+1)\mod M$,
 $\text{\texttt{ADD}}(a,b)=(a+b)\mod M$,
 $\text{\texttt{SUB}}(a,b)=(a-b)\mod M$,
 $\text{\texttt{DEC}}(a,b)=(a-1)\mod M$,
 $\text{\texttt{LESS-THAN}}(a,b)=[a<b]$,
 $\text{\texttt{LESS-OR-EQUAL-THAN}}(a,b)=[a \le b]$,
 $\text{\texttt{EQUALITY-TEST}}(a,b)=[a = b]$,
 $\text{\texttt{MIN}}(a,b)=min(a,b)$,
 $\text{\texttt{MAX}}(a,b)=max(a,b)$,
 \texttt{WRITE} (described in Sec.~\ref{sec:memory}).

We also considered settings in which the module sequence is repeated many times, e.g.
there are $28$ modules, where modules number $1.$ and $15.$ are \texttt{READ},
modules number $2.$ and $16.$ are \texttt{ZERO} and so on.
The number of repetitions is a hyperparameter.

\subsection{Results}
\label{experiments_results}

\begin{table}[t]
\begin{center}
\small
\begin{tabular}{|c|c|c|c|c|}
\hline
\textbf{Task} & \textbf{Train Complexity} & \textbf{Train error} & \textbf{Generalization} & \textbf{Discretization} \\
\hline
Access & $len(A) \le 20$ & 0 & perfect & perfect \\
Increment & $len(A) \le 15$ & 0 & perfect & perfect \\
Copy & $len(A) \le 15$ & 0 & perfect & perfect \\
Reverse & $len(A) \le 15$ & 0 & perfect& perfect \\
Swap & $len(A) \le 20$ & 0 & perfect & perfect \\ \hline
Permutation & $len(A) \le 6$ & 0 & almost perfect & perfect \\
ListK & $len(list) \le 10$ & 0 & strong & hurts performance \\
ListSearch & $len(list) \le 6$ & 0 & weak & hurts performance \\
Merge & $len(A) + len(B) \le 10$ & $1\%$ & weak & hurts performance \\
WalkBST & $size(tree) \le 10$ & $0.3\%$ & strong & hurts performance \\
\hline
\end{tabular}
\end{center}
\caption{Results of the experiments.
The \textbf{perfect} generalization error means that the tested problem had error $0$ for complexity up to $50$.
Exact generalization errors are presented in Fig.~\ref{fig:generalization}
The \textbf{perfect} discretization means that the discretized version of the model
produced exactly the same outputs as the original model on all test cases.}
\label{table_complexity}
\end{table}

\begin{figure}[h]
\setkeys{Gin}{width=0.6\textwidth}
\centering
\includegraphics{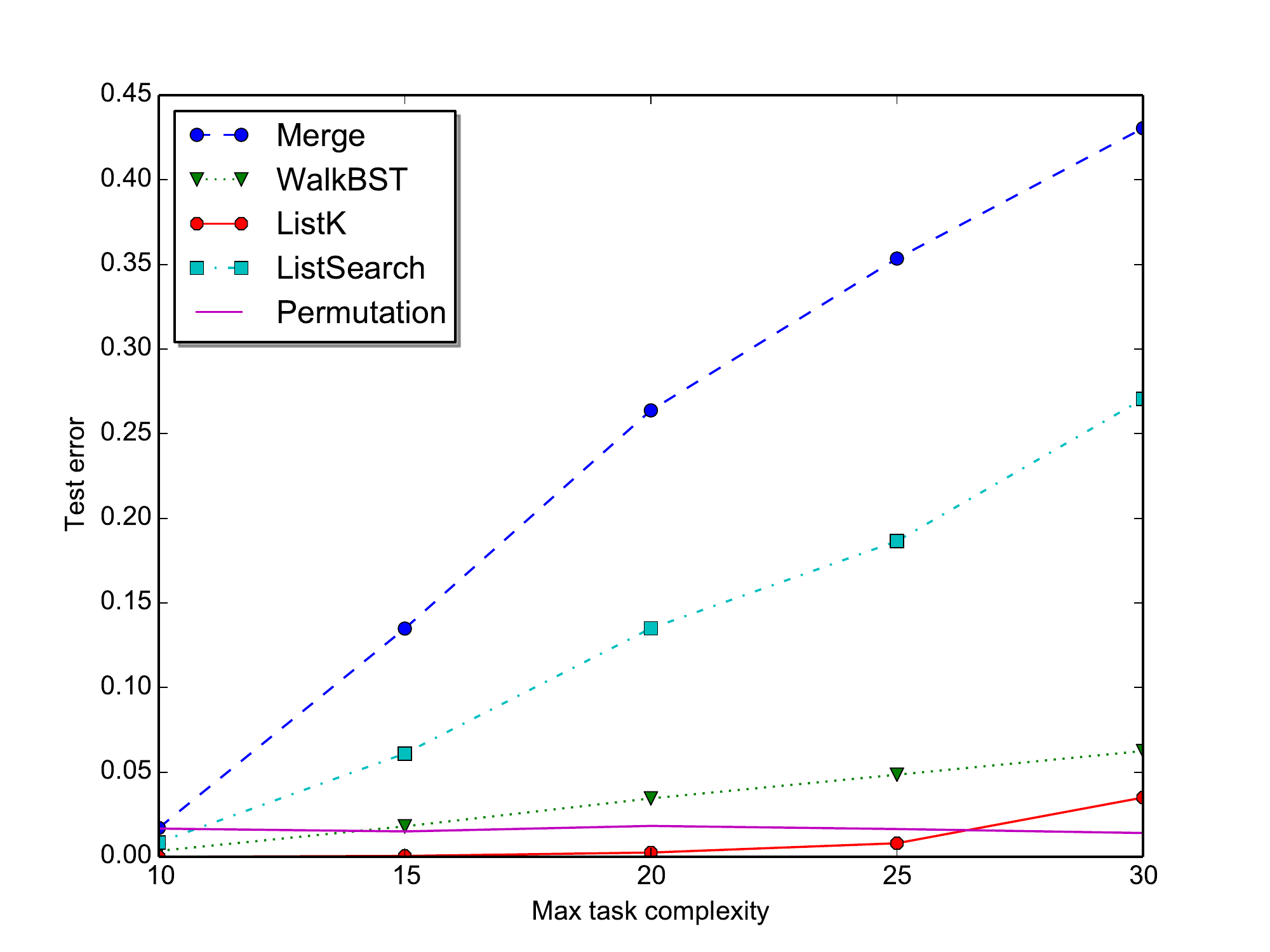}
\caption{Generalization errors for hard tasks. The \textbf{Permutation} and \textbf{ListSearch} problems
were trained only up to complexity $6$.
The remaining problems were trained up to complexity $10$.
The horizontal axis denotes the \emph{maximal} task complexity, i.e., $x=20$ denotes results with complexity sampled uniformly from the interval $[1,20]$.}
\label{fig:generalization}
\end{figure}

Overall, we were able to find parameters that achieved an error 0 for all problems except \textbf{Merge} and \textbf{WalkBST}
(where we got an error of $\le 1\%$). As described in \ref{sec:tasks}, our metric is an accuracy
on the memory cells that should be modified. To compute it, we take the continuous memory state produced
by our network, then discretize it (every cell will contain the value with the highest probability),
and finally compare with the expected output.
The results of the experiments are summarized in Table~\ref{table_complexity}.

Below we describe our results on all $10$ tasks in more detail. We divide them
into $2$ categories: "easy" and "hard" tasks. Easy tasks is a category of tasks
that achieved low error scores for many sets of parameters and we did not have to
spend much time trying to tune them. First $5$ problems from our task list belong
to this category. Hard tasks, on the other hand, are problems that often trained to low error rate only in a very small number of cases, eg. $1$ out of
$100$.

\subsubsection{Easy tasks}
This category includes the following problems: \textbf{Access},
\textbf{Increment}, \textbf{Copy}, \textbf{Reverse}, \textbf{Swap}.  For all of
them we were able to find many sets of hyperparameters that achieved error 0, or
close to it without much effort.

We also tested how those solutions generalize to longer input sequences.
To do this, for every problem we selected a model that achieved error $0$ during the training,
and tested it on inputs with lengths up to $50$\footnote{Unfortunately we could not test for lengths longer than $50$ due to the memory restrictions.}.
To perform these tests we also increased the memory size and the number of allowed timesteps.

In all cases the model solved the problem perfectly, what shows that it
generalizes not only to longer input sequences, but also to different memory
sizes and numbers of allowed timesteps.  Moreover, the discretized version of
the model (see Sec.~\ref{sec:discrete} for details) also solves all the
problems perfectly. These results show that the NRAM model naturally learns
``algorithmic'' solutions, which generalize well.

We were also interested if the found solutions generalize to sequences of
arbitrary length.  It is easiest to verify in the case of a discretized model
with a feedforward controller. That is because then circuits outputted by the
controller depend solely on the values of registers, which are integers.  We
manually analysed circuits for problems \textbf{Copy} and \textbf{Increment}
and verified that found solutions generalize to inputs of arbitrary length,
assuming that the number of allowed timesteps is appropriate.

\begin{table}[t!]
\footnotesize

\input{memory_table}

\caption{\small State of memory and registers for the \textbf{Copy}
  problem at the start of every timestep. We also show the arguments
  given to the \texttt{READ} and \texttt{WRITE} functions in each
  timestep. The argument ``p:'' represents the source/destination
  address and ``a:'' represents the value to be written (for
  \texttt{WRITE}).  The value $6$ at position $0$ in the memory is the
  pointer to the destination array. It is followed by $5$ values (gray
  columns) that should be copied.}

\label{table_copy}
\end{table}

\begin{wrapfigure}[19]{r}{0.5\textwidth}
\vspace{-1.7cm}
\centering
\includegraphics[scale=0.45]{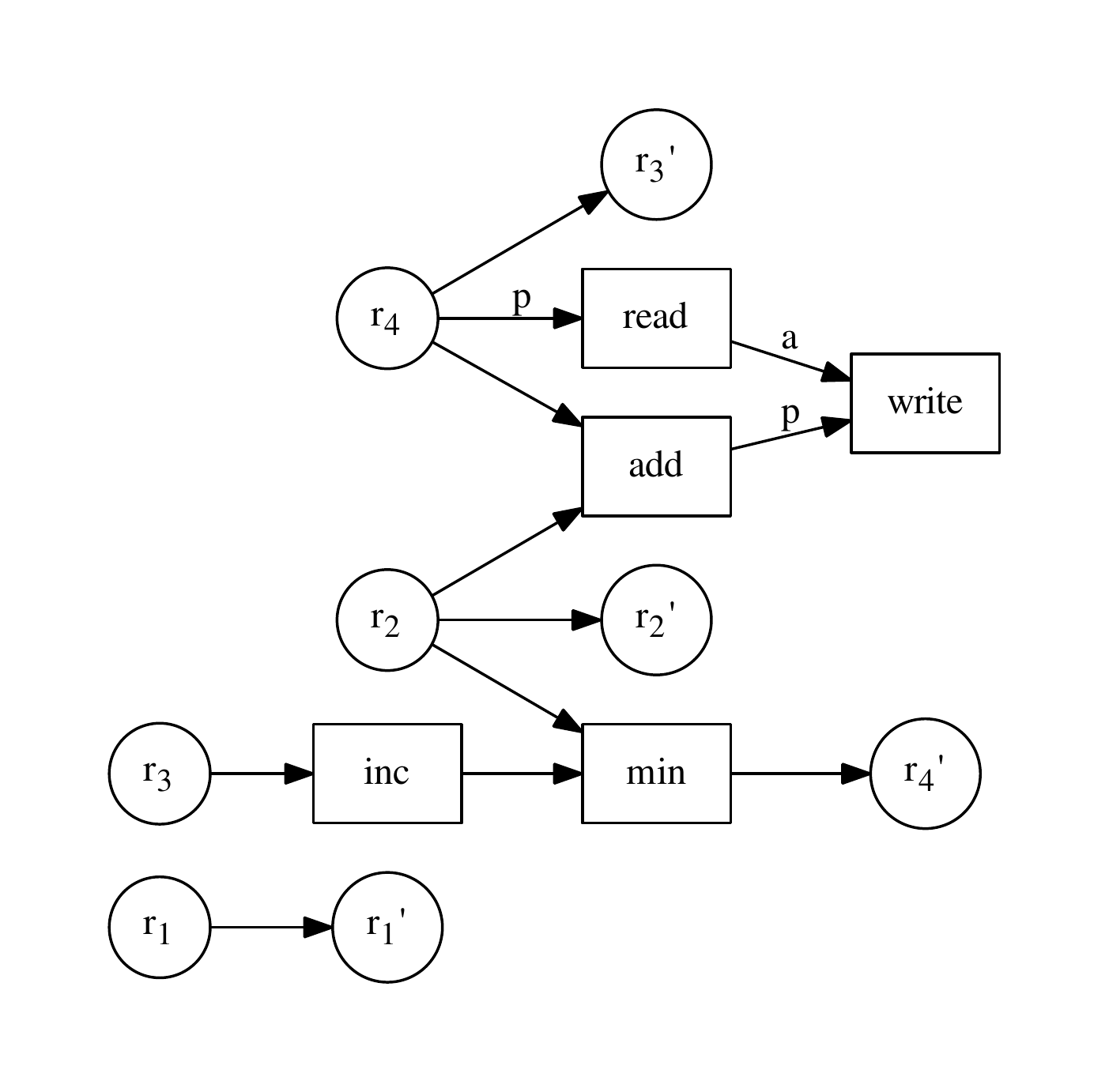}
\caption{The circuit generated at every timestep $\ge 2$.
The values of the pointer ($p$) for \texttt{READ}, \texttt{WRITE} and the value to be written ($a$) for \texttt{WRITE} are presented in Table \ref{table_copy}.
The modules whose outputs are not used were removed from the picture.}
\label{fig:circuit_2}
\end{wrapfigure}

\subsubsection{Hard tasks}

This category includes: \textbf{Permutation}, \textbf{ListK},
\textbf{ListSearch}, \textbf{Merge} and \textbf{WalkBST}.
For all of them we had to perform an extensive random search to find a good set of hyperparameters.
Usually, most of the parameter
combinations were stuck on the starting curriculum level with a high error
of $50\% - 70\%$.
For the first $3$ tasks we managed to train the network to achieve error $0$.
For \textbf{WalkBST} and \textbf{Merge} the training errors were $0.3$\% and $1$\% respectively.
For training those problems we had to introduce additional techniques described in
Sec.~\ref{training_procedure}.

For \textbf{Permutation}, \textbf{ListK} and \textbf{WalkBST} our model generalizes very well
and achieves low error rates on inputs at least twice longer than the ones seen during the training.
The exact generalization errors are shown in Fig.~\ref{fig:generalization}.

The only hard problem on which our model \mbox{discretizes} well is
\textbf{Permutation} --- on this task the discretized version of the model
produces exactly the same outputs as the original model on all cases tested.
For the remaining four problems the discretized version of the models perform
very poorly (error rates $\ge 70\%$).  We believe that better results may be
obtained by using some techniques encouraging discretization during the training
\footnote{One could for example add at later stages of training a penalty
proportional to the entropy of the intermediate values of registers/memory.}.

We noticed that the training procedure is very unstable and the error
often raises from a few percents to e.g. $70\%$ in just one epoch. Moreover,
even if we use the best found set of hyperparameters, the percent of random
seeds that converges to error $0$ was usually equal about $11\%$. We observed
that the percent of converging seeds is much lower if we do not add noise to the
gradient --- in this case only about $1\%$ of seeds converge.

\subsection{Comparison to existing models}

A comparison to other models is challenging because we are the first to consider
problems with pointers. The NTM can solve tasks like \textbf{Copy} or
\textbf{Reverse}, but it suffers from the inability to naturally store a pointer
to a fixed location in the memory. This makes it unlikely that it could solve tasks
such as {\bf ListK, ListSearch} or {\bf WalkBST} since the pointers used in
these tasks refer to absolute positions.

What distinguishes our model from most of the previous attempts 
(including NTMs, Memory Networks, Pointer Networks) is the lack of 
content-based addressing. It was a deliberate design decision, since this kind
of addressing inherently slows down the memory access. In contrast, our model
 --- if discretized --- can access the memory in a constant time.

The NRAM is also the first model that we are aware of employing a differentiable
mechanism for deciding when to finish the computation.

\subsection{Exemplary execution}

We present one example execution of our model for the problem \textbf{Copy}. 
For the example, we use a very small model with $12$ memory
cells, $4$ registers and the standard set of $14$ modules. The controller for this
model is a feedforward network, and we run it for $11$ timesteps.
Table \ref{table_copy} contains, for every timestep, the state of the memory
and registers at the begin of the timestep. 

The model can execute different circuits at different timesteps. In particular, we
observed that the first circuit is slightly different from the rest, since it
needs to handle the initialization.  Starting from the second step all generated
circuits are the same. We present this circuit in
Fig.~\ref{fig:circuit_2}.
The register $r_2$ is constant and keeps the offset between the destination array and the source array ($6-1=5$ in
this case). The register $r_3$ is responsible for incrementing the pointer in
the source array. Its value is copied to $r_4$\footnote{In our case $r_3 < r_2$, so the \texttt{MIN} module always outputs the value $r_3+1$. It is not satisfied in the last timestep, but then the array is already copied.},
the register used by the
\texttt{READ} module. For the \texttt{WRITE} module, it also uses $r_4$ which is
shifted by $r_2$.
The register $r_1$ is unused. This solution generalizes to sequences of arbitrary length.

%% file: memory_table.tex
\definecolor{Gray}{gray}{0.8}
\newcolumntype{g}{>{\columncolor{Gray}}c}

\begin{center}
\begin{tabular}{|c|cgggggcccccc|cccc|c|c|}
\hline
\textbf{Step} & 0 & 1 & 2 & 3 & 4 & 5 & 6 & 7 & 8 & 9 & 10 & 11 & $r_1$ & $r_2$ & $r_3$ & $r_4$ & \texttt{READ} & \texttt{WRITE} \\
\hline
  
1 & 6 & 2 & 10 & 6 & 8 & 9 & 0 & 0 & 0 & 0 & 0 & 0 & 0 & 0 & 0 & 0& p:0& p:0 a:6 \\
2 & 6 & 2 & 10 & 6 & 8 & 9 & 0 & 0 & 0 & 0 & 0 & 0 & 0 & 5 & 0 & 1& p:1& p:6 a:2 \\
3 & 6 & 2 & 10 & 6 & 8 & 9 & 2 & 0 & 0 & 0 & 0 & 0 & 0 & 5 & 1 & 1& p:1& p:6 a:2 \\
4 & 6 & 2 & 10 & 6 & 8 & 9 & 2 & 0 & 0 & 0 & 0 & 0 & 0 & 5 & 1 & 2& p:2& p:7 a:10 \\
5 & 6 & 2 & 10 & 6 & 8 & 9 & 2 & 10 & 0 & 0 & 0 & 0 & 0 & 5 & 2 & 2& p:2& p:7 a:10 \\
6 & 6 & 2 & 10 & 6 & 8 & 9 & 2 & 10 & 0 & 0 & 0 & 0 & 0 & 5 & 2 & 3& p:3& p:8 a:6 \\
7 & 6 & 2 & 10 & 6 & 8 & 9 & 2 & 10 & 6 & 0 & 0 & 0 & 0 & 5 & 3 & 3& p:3& p:8 a:6 \\
8 & 6 & 2 & 10 & 6 & 8 & 9 & 2 & 10 & 6 & 0 & 0 & 0 & 0 & 5 & 3 & 4& p:4& p:9 a:8 \\
9 & 6 & 2 & 10 & 6 & 8 & 9 & 2 & 10 & 6 & 8 & 0 & 0 & 0 & 5 & 4 & 4& p:4& p:9 a:8 \\
10 & 6 & 2 & 10 & 6 & 8 & 9 & 2 & 10 & 6 & 8 & 0 & 0 & 0 & 5 & 4 & 5& p:5& p:10 a:9 \\
11 & 6 & 2 & 10 & 6 & 8 & 9 & 2 & 10 & 6 & 8 & 9 & 0 & 0 & 5 & 5 & 5& p:5& p:10 a:9 \\

\hline
\end{tabular}
\end{center}

%% file: conclusions.tex
\section{Conclusions}

In this paper we presented the Neural Random-Access Machine, which can
learn to solve problems that require explicit manipulation and
dereferencing of pointers.  

We showed that this model can learn to solve a number of algorithmic problems
and generalize well to inputs longer than ones seen during the training. In
particular, for some problems it generalizes to inputs of arbitrary length.

However, we noticed that the optimization problem resulting from the backpropagating through
the execution trace of the program is very challenging for standard optimization techniques. 
It seems likely that a method that can search in an easier ``abstract'' space would be
more effective at solving such problems.

%% file: task_descriptions.tex
\section{Detailed tasks descriptions}
\label{appendix:tasks}

In this section we describe in details the memory layout of inputs and outputs
for the tasks used in our experiments. In all descriptions below, big letters represent
arrays and small letters represents pointers. $NULL$ denotes the value $0$ and
is used to mark the end of an array or a missing next element in a list or
a binary tree.

\begin{enumerate}

\item \textbf{Access} Given a value $k$ and an array $A$, return $A[k]$. Input is
given as $k, A[0], .., A[n-1], NULL$ and the network should replace the first memory cell
with $A[k]$.

\item \textbf{Increment} Given an array $A$, increment all its elements by 1.
Input is given as $A[0], ..., A[n-1], NULL$ and the expected output is
$A[0] + 1, ..., A[n-1] + 1$.

\item \textbf{Copy} Given an array and a pointer to the destination, copy all
elements from the array to the given location. Input is given as $p, A[0], ...,
A[n-1]$ where $p$ points to one element after $A[n-1]$. The expected output is
$A[0], ..., A[n-1]$ at positions $p, ..., p+n-1$ respectively.

\item \textbf{Reverse} Given an array and a pointer to the destination, copy
all elements from the array in reversed order.  Input is given as $p, A[0],
..., A[n-1]$ where $p$ points one element after $A[n-1]$. The expected output
is $A[n-1], ..., A[0]$ at positions $p, ..., p+n-1$ respectively.

\item \textbf{Swap} Given two pointers $p$, $q$ and an array $A$, swap elements
$A[p]$ and $A[q]$.  Input is given as $p, q, A[0], .., A[p], ..., A[q], ...,
A[n-1], 0$. The expected modified array $A$ is: $A[0],
..., A[q], ..., A[p], ..., A[n-1]$.

\item \textbf{Permutation} Given two arrays of $n$ elements: $P$ (contains
a permutation of numbers ${0,\ldots,n-1}$) and $A$ (contains random elements), permutate $A$
according to $P$. Input is given as $a, P[0], ..., P[n-1], A[0], ..., A[n-1]$,
where $a$ is a pointer to the array $A$. The expected output is
$A[P[0]], ..., A[P[n-1]]$, which should override the array $P$.

\item \textbf{ListK} Given a pointer to the head of a linked list and a number $k$,
find the value of the $k$-th element on the list.  List nodes are represented as two adjacent
memory cells: a pointer to the next node and a value. Elements are in random
locations in the memory, so that the network needs to follow the pointers to
find the correct element.  Input is given as: $head, k, out, ...$ where head is a pointer to the first node on the list, $k$
indicates how many hops are needed and $out$ is a cell where the output should be put.

\item \textbf{ListSearch} Given a pointer to the head of a linked list and a value $v$ to find
return a pointer to the first node on the list with the value $v$.
The list is placed in memory in the same way as in the task \textbf{ListK}.
We fill empty memory with ``trash'' values
to prevent the network from ``cheating'' and just iterating over the whole memory.

\item \textbf{Merge}
Given pointers to $2$ sorted arrays $A$ and $B$, and the pointer to the output
$o$, merge the two arrays into one sorted array. The input is given as: $a, b,
o, A[0], .., A[n-1], G, B[0], ..., B[m-1],G$, where $G$ is a special guardian
value, $a$ and $b$ point to the first elements of arrays $A$ and $B$ respectively, and
$o$ points to the address after the second $G$. The $n + m$ element should be
written in correct order starting from position $o$.

\item \textbf{WalkBST} 
Given a pointer to the root of a Binary Search Tree, and a path to be
traversed, return the element at the end of the path. The BST nodes are
represented as tripes ($v$, $l$, $r$), where $v$ is the value, and $l$, $r$ are
pointers to the left/right child. The triples are placed randomly in the
memory. Input is given as $root, out, d_1, d_2, ..., d_k, NULL, ...$, where
$root$ points to the root node and $out$ is a slot for the output. The
sequence $d_1 ... d_k, d_i \in \{0, 1\}$ represents the path to be traversed:
$d_i = 0$ means that the network should go to the left child, $d_i = 1$
represents going to the right child.

\end{enumerate}

%% file: appendix_curriculum.tex
\section{Details of curriculum training}\label{appendix:curriculum}
As noticed in several papers \citep{curriculum,lte}, curriculum
learning is crucial for training deep networks on very complicated
problems. We followed the curriculum learning schedule from
\citet{lte} without any modifications.

For each of the tasks we have manually defined a sequence of subtasks with
increasing difficulty, where the difficulty is usually measured by the length
of the input sequence.  During training the input-output examples are sampled
from a distribution that is determined by the current difficulty level $D$.
The level is increased (up to some maximal value) whenever the error rate of
the model goes below some threshold.  Moreover, we ensure that successive
increases of $D$ are separated by some number of batches.

In more detail, to generate an input-output example we first sample
a difficulty $d$ from a distribution determined by the current level $D$
and then draw the example with the difficulty $d$.
The procedure for sampling $d$ is the following:
\begin{itemize}
 \item with probability $10\%$: pick $d$ uniformly at random from the set of
 all possible difficulties;
 \item with probability $25\%$: pick $d$ uniformly from $[1,D+e]$, where $e$
 is a sample from a geometric distribution with a success probability $1/2$;
 \item with probability $65\%$: set $d=D+e$, where $e$ is sampled as above.
\end{itemize}

Notice that the above procedure guarantees that every difficulty $d$
can be picked regardless of the current level $D$, which has been shown
to increase performance \cite{lte}.

%% file: appendix_circuits.tex
\section{Example Circuits}\label{appendix:circuits}

Below are presented example circuits generated during training for all simple
tasks (except \textbf{Copy} which was presented in the paper). For modules
\texttt{READ} and \texttt{WRITE}, the value of the first argument (pointer to
the address to be read/written) is marked as $p$. For \texttt{WRITE}, the value
to be written is marked as $a$ and the value returned by this module is always
$0$.  For modules \texttt{LESS-THAN} and \texttt{LESS-OR-EQUAL-THAN} the first
parameter is marked as $x$ and the second one as $y$. Other modules either have
only one parameter or the order of parameters is not important. 

For all tasks below (except \textbf{Increment}), the circuit generated at timestep $1$
is different than circuits generated at steps $\ge 2$, which are the same. This is because
the first circuit needs to handle the initialization. We present only the "main" circuits
generated for timesteps $\ge 2$.

\subsection{Access}
\begin{figure}[h!]
\centering
\includegraphics[scale=0.6]{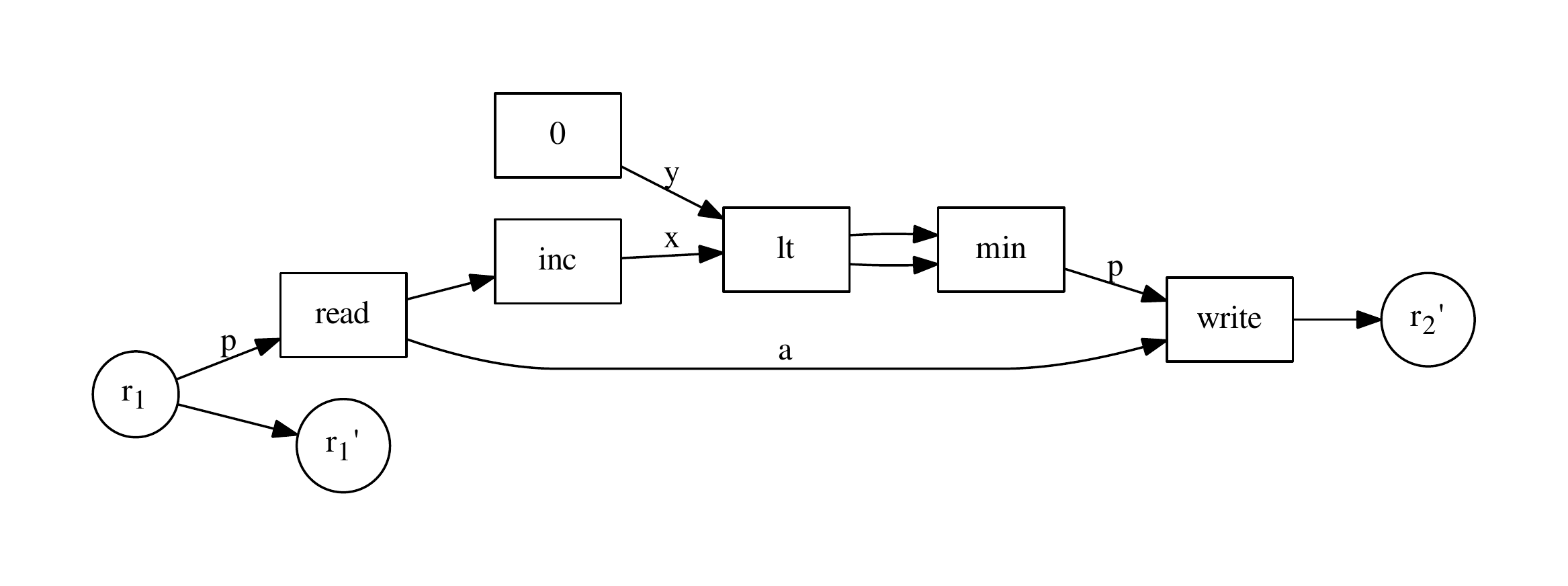}
\caption{The circuit generated at every timestep $\ge 2$ for the task \textbf{Access}.} 
\label{fig:circuit_access}
\end{figure}

\begin{table}[h!]
\footnotesize
\input{memory_table_access}

\caption{Memory for task \textbf{Access}. Only the first memory cell is modified.}
\label{table_memory_access}
\end{table}

\newpage
\subsection{Increment}
\begin{figure}[h!]
\centering
\includegraphics[scale=0.6]{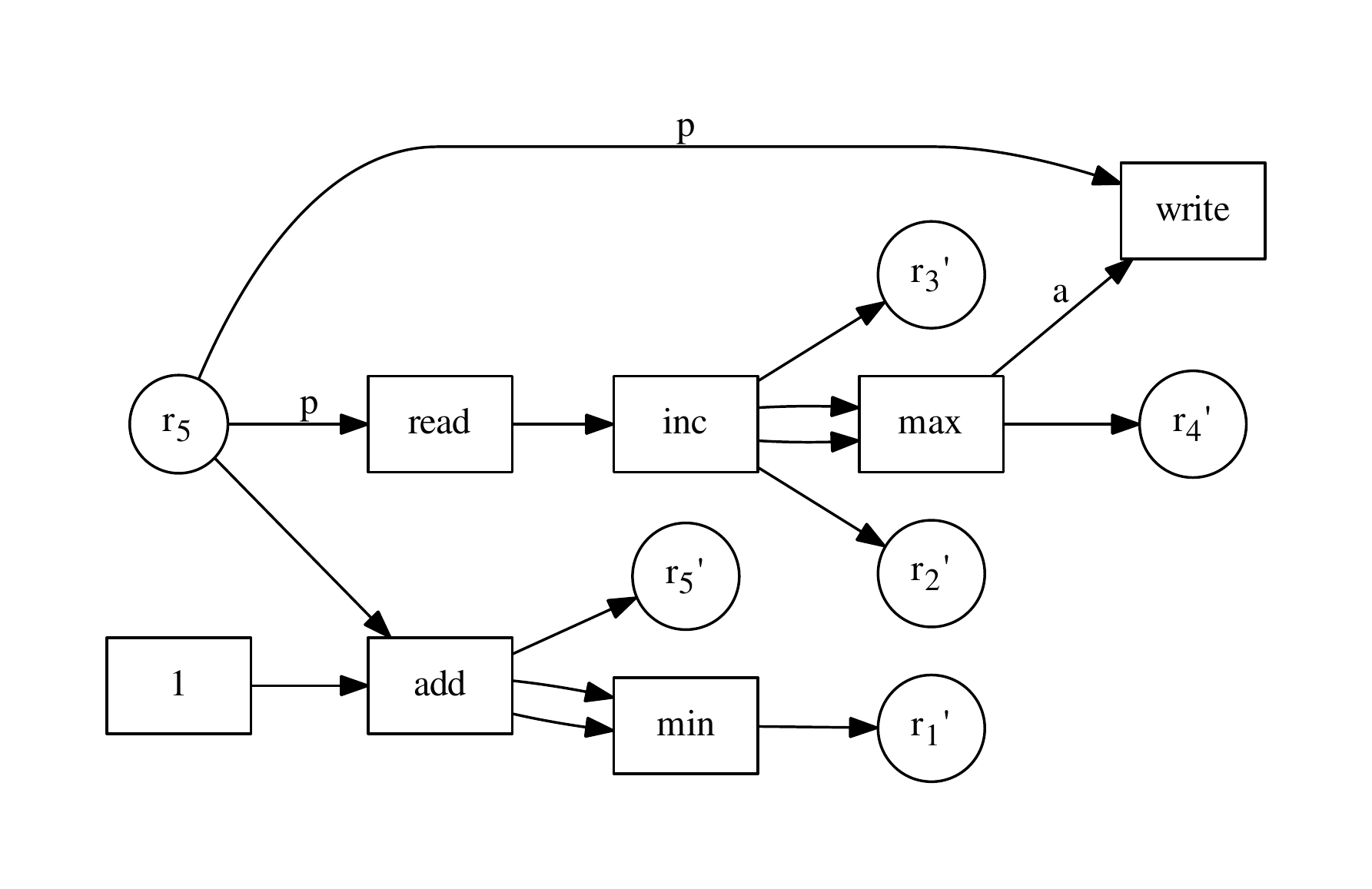}
\caption{The circuit generated at every timestep for the task \textbf{Increment}.}
\label{fig:circuit_increment}
\end{figure}

\begin{table}[h!]
\footnotesize
\input{memory_table_inc}

\caption{Memory for task \textbf{Increment}.}
\label{table_memory_inc}
\end{table}

\newpage
\subsection{Reverse}
\begin{figure}[h!]
\centering
\includegraphics[scale=0.6]{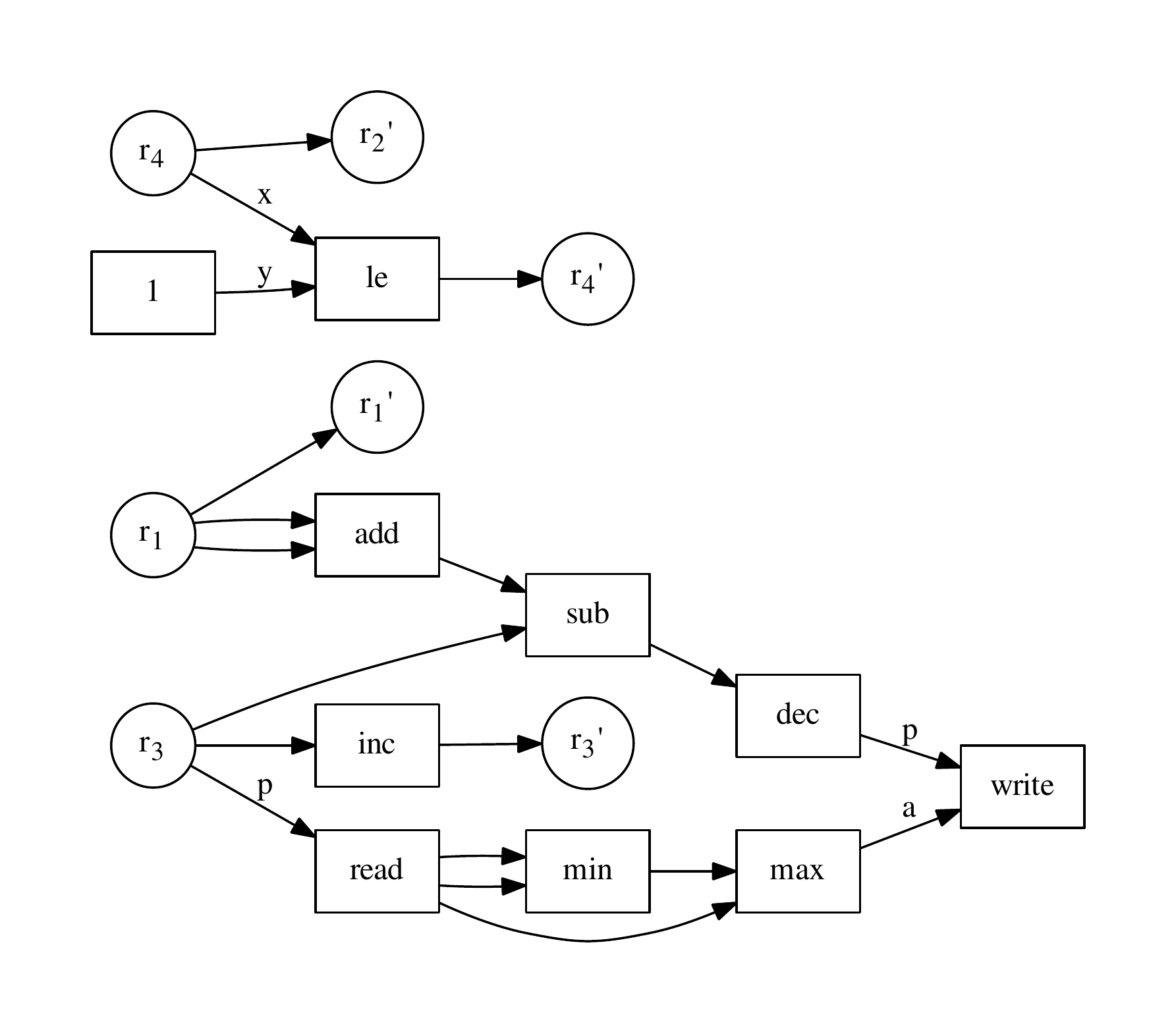}
\caption{The circuit generated at every timestep $\ge 2$ for the task \textbf{Reverse}.}
\label{fig:circuit_reverse}
\end{figure}

\begin{table}[h!]
\footnotesize
\input{memory_table_reverse}

\caption{Memory for task \textbf{Reverse}.}
\label{table_memory_reverse}
\end{table}

\newpage
\subsection{Swap}

For swap we observed that $2$ different circuits are generated, one for even timesteps, one for odd timesteps.
\begin{figure}[h!]
\centering
\includegraphics[scale=0.5]{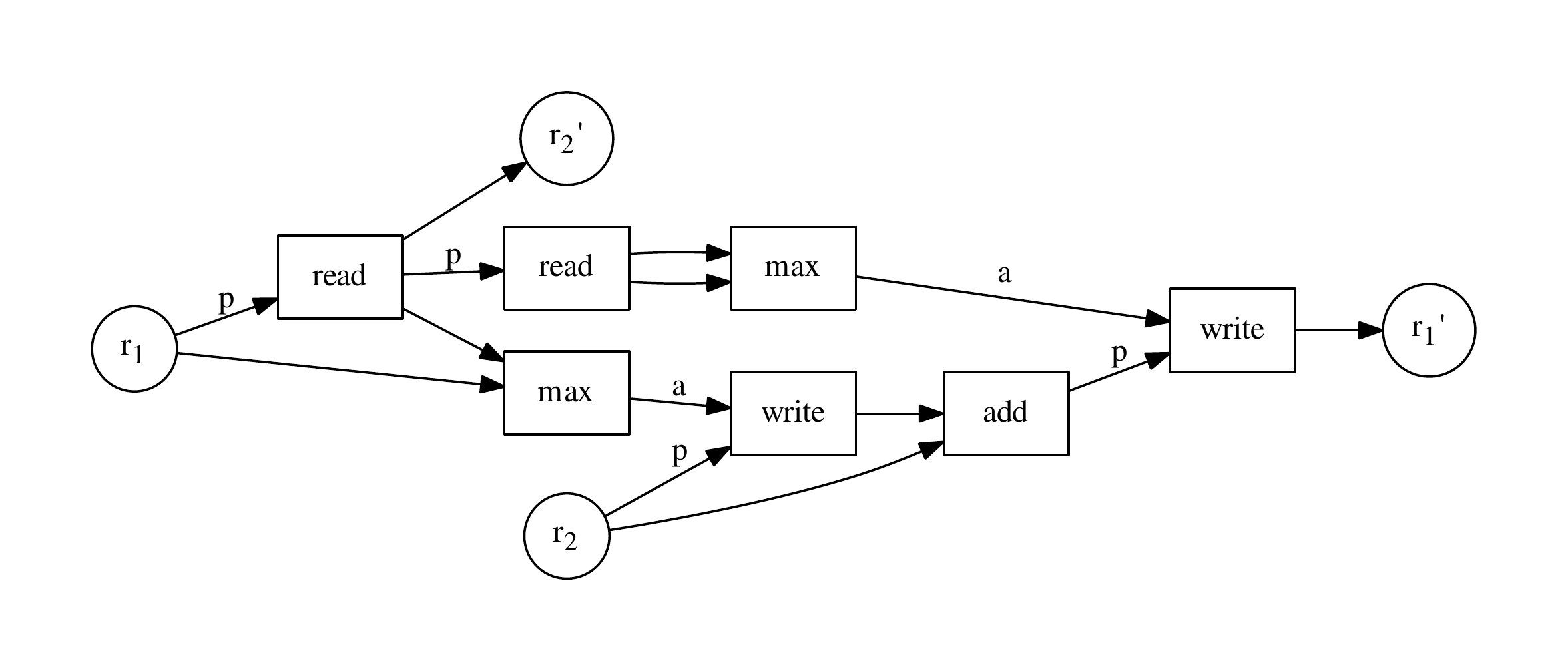}
\caption{The circuit generated at every even timestep for the task \textbf{Swap}.}
\label{fig:circuit_swap2}
\end{figure}

\begin{figure}[h!]
\centering
\includegraphics[scale=0.5]{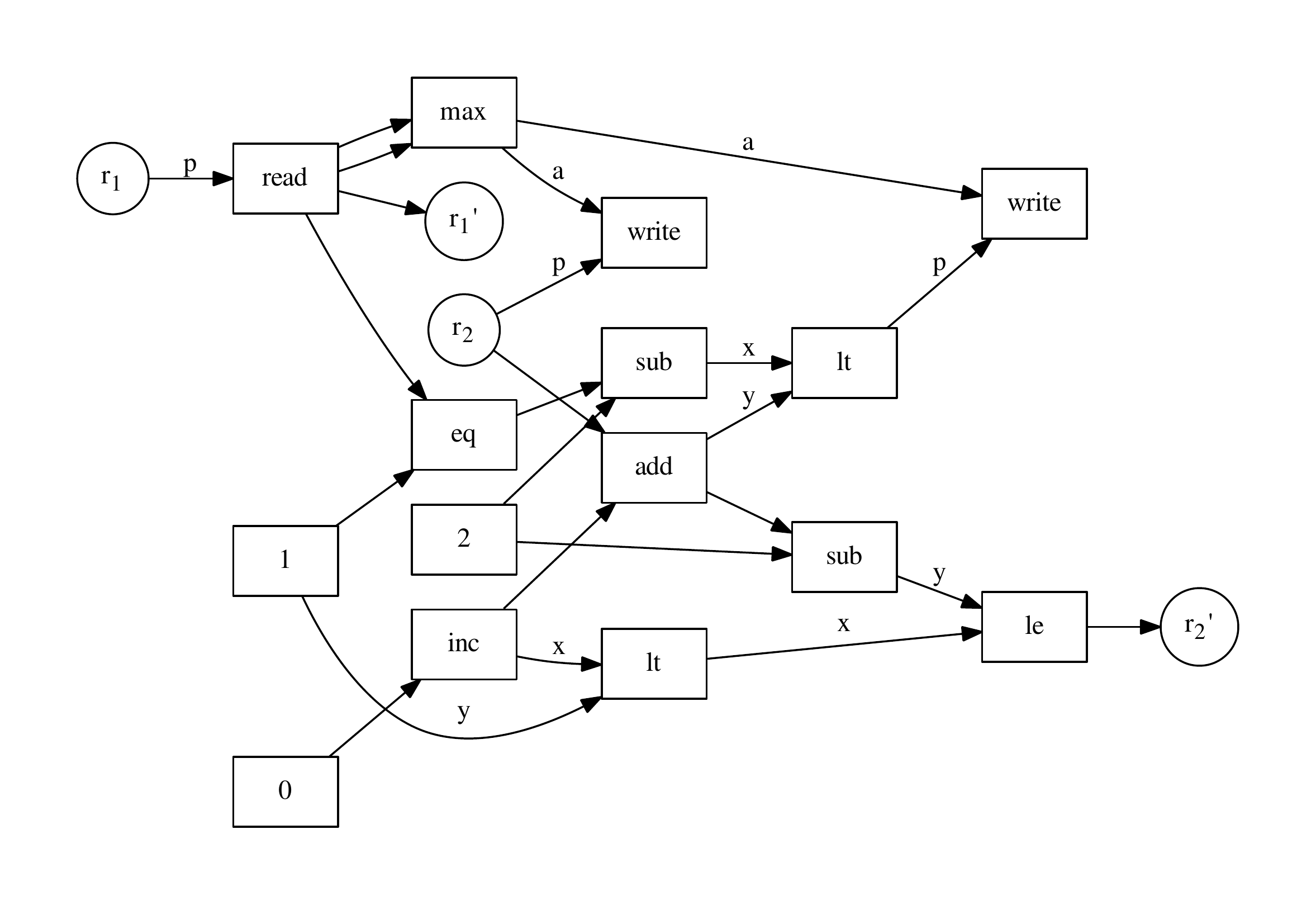}
\caption{The circuit generated at every odd timestep $\ge 3$ for the task \textbf{Swap}.}
\label{fig:circuit_swap3}
\end{figure}

\begin{table}[h!]
\footnotesize
\input{memory_table_swap}

\caption{Memory for task \textbf{Swap}.}
\label{table_memory_swap}
\end{table}

%% file: memory_table_access.tex
\definecolor{Gray}{gray}{0.8}
\newcolumntype{g}{>{\columncolor{Gray}}c}
\begin{center}
\begin{tabular}{|c|cccccccccccccccc|cc|}
\hline
\textbf{Step} & 0 & 1 & 2 & 3 & 4 & 5 & 6 & 7 & 8 & 9 & 10 & 11 & 12 & 13 & 14 & 15 & $r_1$ & $r_2$\\
\hline
1 & 3 & 1 & 12 & 4 & 7 & 12 & 1 & 13 & 8 & 2 & 1 & 3 & 11 & 11 & 12 & 0 & 0 & 0 \\
2 & 3 & 1 & 12 & 4 & 7 & 12 & 1 & 13 & 8 & 2 & 1 & 3 & 11 & 11 & 12 & 0 & 3 & 0 \\
3 & 4 & 1 & 12 & 4 & 7 & 12 & 1 & 13 & 8 & 2 & 1 & 3 & 11 & 11 & 12 & 0 & 3 & 0 \\

\hline
\end{tabular}
\end{center}

%% file: memory_table_inc.tex
\definecolor{Gray}{gray}{0.8}
\newcolumntype{g}{>{\columncolor{Gray}}c}
\begin{center}
\begin{tabular}{|c|cccccccccccccccc|ccccc|}
\hline
\textbf{Step} & 0 & 1 & 2 & 3 & 4 & 5 & 6 & 7 & 8 & 9 & 10 & 11 & 12 & 13 & 14 & 15 & $r_1$ & $r_2$ & $r_3$ & $r_4$ & $r_5$\\
\hline
1 & 1 & 11 & 3 & 8 & 1 & 2 & 9 & 8 & 5 & 3 & 0 & 0 & 0 & 0 & 0 & 0 & 0 & 0 & 0 & 0 & 0 \\
2 & 2 & 11 & 3 & 8 & 1 & 2 & 9 & 8 & 5 & 3 & 0 & 0 & 0 & 0 & 0 & 0 & 1 & 2 & 2 & 2 & 1 \\
3 & 2 & 12 & 3 & 8 & 1 & 2 & 9 & 8 & 5 & 3 & 0 & 0 & 0 & 0 & 0 & 0 & 2 & 12 & 12 & 12 & 2 \\
4 & 2 & 12 & 4 & 8 & 1 & 2 & 9 & 8 & 5 & 3 & 0 & 0 & 0 & 0 & 0 & 0 & 3 & 4 & 4 & 4 & 3 \\
5 & 2 & 12 & 4 & 9 & 1 & 2 & 9 & 8 & 5 & 3 & 0 & 0 & 0 & 0 & 0 & 0 & 4 & 9 & 9 & 9 & 4 \\
6 & 2 & 12 & 4 & 9 & 2 & 2 & 9 & 8 & 5 & 3 & 0 & 0 & 0 & 0 & 0 & 0 & 5 & 2 & 2 & 2 & 5 \\
7 & 2 & 12 & 4 & 9 & 2 & 3 & 9 & 8 & 5 & 3 & 0 & 0 & 0 & 0 & 0 & 0 & 6 & 3 & 3 & 3 & 6 \\
8 & 2 & 12 & 4 & 9 & 2 & 3 & 10 & 8 & 5 & 3 & 0 & 0 & 0 & 0 & 0 & 0 & 7 & 10 & 10 & 10 & 7 \\
9 & 2 & 12 & 4 & 9 & 2 & 3 & 10 & 9 & 5 & 3 & 0 & 0 & 0 & 0 & 0 & 0 & 8 & 9 & 9 & 9 & 8 \\
10 & 2 & 12 & 4 & 9 & 2 & 3 & 10 & 9 & 6 & 3 & 0 & 0 & 0 & 0 & 0 & 0 & 9 & 6 & 6 & 6 & 9 \\
11 & 2 & 12 & 4 & 9 & 2 & 3 & 10 & 9 & 6 & 4 & 0 & 0 & 0 & 0 & 0 & 0 & 10 & 4 & 4 & 4 & 10 \\

\hline
\end{tabular}
\end{center}

%% file: memory_table_reverse.tex
\definecolor{Gray}{gray}{0.8}
\newcolumntype{g}{>{\columncolor{Gray}}c}
\begin{center}
\begin{tabular}{|c|cccccccccccccccc|cccc|}
\hline
\textbf{Step} & 0 & 1 & 2 & 3 & 4 & 5 & 6 & 7 & 8 & 9 & 10 & 11 & 12 & 13 & 14 & 15 & $r_1$ & $r_2$ & $r_3$ & $r_4$\\
\hline
1 & 8 & 8 & 1 & 3 & 5 & 1 & 1 & 2 & 0 & 0 & 0 & 0 & 0 & 0 & 0 & 0 & 0 & 0 & 0 & 0 \\
2 & 8 & 8 & 1 & 3 & 5 & 1 & 1 & 2 & 0 & 0 & 0 & 0 & 0 & 0 & 0 & 0 & 8 & 0 & 1 & 1 \\
3 & 8 & 8 & 1 & 3 & 5 & 1 & 1 & 2 & 0 & 0 & 0 & 0 & 0 & 0 & 8 & 0 & 8 & 1 & 2 & 1 \\
4 & 8 & 8 & 1 & 3 & 5 & 1 & 1 & 2 & 0 & 0 & 0 & 0 & 0 & 1 & 8 & 0 & 8 & 1 & 3 & 1 \\
5 & 8 & 8 & 1 & 3 & 5 & 1 & 1 & 2 & 0 & 0 & 0 & 0 & 3 & 1 & 8 & 0 & 8 & 1 & 4 & 1 \\
6 & 8 & 8 & 1 & 3 & 5 & 1 & 1 & 2 & 0 & 0 & 0 & 5 & 3 & 1 & 8 & 0 & 8 & 1 & 5 & 1 \\
7 & 8 & 8 & 1 & 3 & 5 & 1 & 1 & 2 & 0 & 0 & 1 & 5 & 3 & 1 & 8 & 0 & 8 & 1 & 6 & 1 \\
8 & 8 & 8 & 1 & 3 & 5 & 1 & 1 & 2 & 0 & 1 & 1 & 5 & 3 & 1 & 8 & 0 & 8 & 1 & 7 & 1 \\
9 & 8 & 8 & 1 & 3 & 5 & 1 & 1 & 2 & 2 & 1 & 1 & 5 & 3 & 1 & 8 & 0 & 8 & 1 & 8 & 1 \\
10 & 8 & 8 & 1 & 3 & 5 & 1 & 1 & 2 & 2 & 1 & 1 & 5 & 3 & 1 & 8 & 0 & 8 & 1 & 9 & 1 \\

\hline
\end{tabular}
\end{center}

%% file: memory_table_swap.tex
\definecolor{Gray}{gray}{0.8}
\newcolumntype{g}{>{\columncolor{Gray}}c}
\begin{center}
\begin{tabular}{|c|cccccccccccccccc|cc|}
\hline
\textbf{Step} & 0 & 1 & 2 & 3 & 4 & 5 & 6 & 7 & 8 & 9 & 10 & 11 & 12 & 13 & 14 & 15 & $r_1$ & $r_2$\\
\hline
1 & 4 & 13 & 6 & 10 & 5 & 4 & 6 & 3 & 7 & 1 & 1 & 11 & 13 & 12 & 0 & 0 & 0 & 0 \\
2 & 5 & 13 & 6 & 10 & 5 & 4 & 6 & 3 & 7 & 1 & 1 & 11 & 13 & 12 & 0 & 0 & 1 & 4 \\
3 & 5 & 13 & 6 & 10 & 12 & 4 & 6 & 3 & 7 & 1 & 1 & 11 & 13 & 12 & 0 & 0 & 0 & 13 \\
4 & 5 & 13 & 6 & 10 & 12 & 4 & 6 & 3 & 7 & 1 & 1 & 11 & 13 & 5 & 0 & 0 & 5 & 1 \\

\hline
\end{tabular}
\end{center}

%% file: paper.bbl
\begin{thebibliography}{20}
\providecommand{\natexlab}[1]{#1}
\providecommand{\url}[1]{\texttt{#1}}
\expandafter\ifx\csname urlstyle\endcsname\relax
  \providecommand{\doi}[1]{doi: #1}\else
  \providecommand{\doi}{doi: \begingroup \urlstyle{rm}\Url}\fi

\bibitem[Bahdanau et~al.(2014)Bahdanau, Cho, and Bengio]{bahdanau2014}
Bahdanau, Dzmitry, Cho, Kyunghyun, and Bengio, Yoshua.
\newblock Neural machine translation by jointly learning to align and
  translate.
\newblock \emph{arXiv preprint arXiv:1409.0473}, 2014.

\bibitem[Bengio et~al.(1994)Bengio, Simard, and Frasconi]{bengio}
Bengio, Yoshua, Simard, Patrice, and Frasconi, Paolo.
\newblock Learning long-term dependencies with gradient descent is difficult.
\newblock \emph{Neural Networks, IEEE Transactions on}, 5\penalty0
  (2):\penalty0 157--166, 1994.

\bibitem[Bengio et~al.(2009)Bengio, Louradour, Collobert, and
  Weston]{curriculum}
Bengio, Yoshua, Louradour, J{\'e}r{\^o}me, Collobert, Ronan, and Weston, Jason.
\newblock Curriculum learning.
\newblock In \emph{Proceedings of the 26th annual international conference on
  machine learning}, pp.\  41--48. ACM, 2009.

\bibitem[Chan et~al.(2015)Chan, Jaitly, Le, and Vinyals]{las}
Chan, William, Jaitly, Navdeep, Le, Quoc~V, and Vinyals, Oriol.
\newblock Listen, attend and spell.
\newblock \emph{arXiv preprint arXiv:1508.01211}, 2015.

\bibitem[Graves et~al.(2014)Graves, Wayne, and Danihelka]{ntm}
Graves, Alex, Wayne, Greg, and Danihelka, Ivo.
\newblock Neural turing machines.
\newblock \emph{arXiv preprint arXiv:1410.5401}, 2014.

\bibitem[Grefenstette et~al.(2015)Grefenstette, Hermann, Suleyman, and
  Blunsom]{stack2}
Grefenstette, Edward, Hermann, Karl~Moritz, Suleyman, Mustafa, and Blunsom,
  Phil.
\newblock Learning to transduce with unbounded memory.
\newblock \emph{arXiv preprint arXiv:1506.02516}, 2015.

\bibitem[Hochreiter \& Schmidhuber(1997)Hochreiter and Schmidhuber]{lstm}
Hochreiter, Sepp and Schmidhuber, J{\"u}rgen.
\newblock Long short-term memory.
\newblock \emph{Neural computation}, 9\penalty0 (8):\penalty0 1735--1780, 1997.

\bibitem[Joulin \& Mikolov(2015)Joulin and Mikolov]{stack-rnn}
Joulin, Armand and Mikolov, Tomas.
\newblock Inferring algorithmic patterns with stack-augmented recurrent nets.
\newblock \emph{arXiv preprint arXiv:1503.01007}, 2015.

\bibitem[Kalchbrenner et~al.(2015)Kalchbrenner, Danihelka, and Graves]{grid}
Kalchbrenner, Nal, Danihelka, Ivo, and Graves, Alex.
\newblock Grid long short-term memory.
\newblock \emph{arXiv preprint arXiv:1507.01526}, 2015.

\bibitem[Kingma \& Ba(2014)Kingma and Ba]{kingma2014adam}
Kingma, Diederik and Ba, Jimmy.
\newblock Adam: A method for stochastic optimization.
\newblock \emph{arXiv preprint arXiv:1412.6980}, 2014.

\bibitem[Luong et~al.(2015)Luong, Pham, and Manning]{thang-mt}
Luong, Minh-Thang, Pham, Hieu, and Manning, Christopher~D.
\newblock Effective approaches to attention-based neural machine translation.
\newblock \emph{arXiv preprint arXiv:1508.04025}, 2015.

\bibitem[Nair \& Hinton(2010)Nair and Hinton]{relu}
Nair, Vinod and Hinton, Geoffrey~E.
\newblock Rectified linear units improve restricted boltzmann machines.
\newblock In \emph{Proceedings of the 27th International Conference on Machine
  Learning (ICML-10)}, pp.\  807--814, 2010.

\bibitem[Neelakantan et~al.(2015)Neelakantan, Vilnis, Le, Sutskever, Kaiser,
  Kurach, and Martens]{neelakantan2015adding}
Neelakantan, Arvind, Vilnis, Luke, Le, Quoc~V, Sutskever, Ilya, Kaiser, Lukasz,
  Kurach, Karol, and Martens, James.
\newblock Adding gradient noise improves learning for very deep networks.
\newblock \emph{arXiv preprint arXiv:1511.06807}, 2015.

\bibitem[Solomonoff(1964)]{solomonoff}
Solomonoff, Ray~J.
\newblock A formal theory of inductive inference. part i.
\newblock \emph{Information and control}, 7\penalty0 (1):\penalty0 1--22, 1964.

\bibitem[Sukhbaatar et~al.(2015)Sukhbaatar, Szlam, Weston, and Fergus]{memnet2}
Sukhbaatar, Sainbayar, Szlam, Arthur, Weston, Jason, and Fergus, Rob.
\newblock End-to-end memory networks.
\newblock \emph{arXiv preprint arXiv:1503.08895}, 2015.

\bibitem[Vinyals et~al.(2014)Vinyals, Kaiser, Koo, Petrov, Sutskever, and
  Hinton]{parsing}
Vinyals, Oriol, Kaiser, Lukasz, Koo, Terry, Petrov, Slav, Sutskever, Ilya, and
  Hinton, Geoffrey.
\newblock Grammar as a foreign language.
\newblock \emph{arXiv preprint arXiv:1412.7449}, 2014.

\bibitem[Vinyals et~al.(2015)Vinyals, Fortunato, and
  Jaitly]{vinyals2015pointer}
Vinyals, Oriol, Fortunato, Meire, and Jaitly, Navdeep.
\newblock Pointer networks.
\newblock \emph{arXiv preprint arXiv:1506.03134}, 2015.

\bibitem[Weston et~al.(2014)Weston, Chopra, and Bordes]{memnet1}
Weston, Jason, Chopra, Sumit, and Bordes, Antoine.
\newblock Memory networks.
\newblock \emph{arXiv preprint arXiv:1410.3916}, 2014.

\bibitem[Zaremba \& Sutskever(2014)Zaremba and Sutskever]{lte}
Zaremba, Wojciech and Sutskever, Ilya.
\newblock Learning to execute.
\newblock \emph{arXiv preprint arXiv:1410.4615}, 2014.

\bibitem[Zaremba \& Sutskever(2015)Zaremba and Sutskever]{rl-ntm}
Zaremba, Wojciech and Sutskever, Ilya.
\newblock Reinforcement learning neural turing machines.
\newblock \emph{arXiv preprint arXiv:1505.00521}, 2015.

\end{thebibliography}
